\documentclass{article}

\usepackage{graphicx} 
\usepackage{subfigure} 
\usepackage{booktabs}

\usepackage[square,sort,comma,numbers]{natbib}

\usepackage{algorithm}
\usepackage{algorithmic}

\usepackage{hyperref}

\usepackage[accepted]{icml2013}

\icmltitlerunning{Convolutional Neural Networks learn compact local image descriptors}

\begin{document} 

\twocolumn[
\icmltitle{Convolutional Neural Networks learn compact local image descriptors}

\icmlauthor{Christian Osendorfer}{osendorf@in.tum.de}
\icmlauthor{Justin Bayer}{bayerj@in.tum.de}
\icmlauthor{Patrick van der Smagt}{smagt@in.tum.de}
\icmladdress{Technische Universit\"{a}t M\"{u}nchen\\
85748 Garching, Germany}

\vskip 0.3in
]

\begin{abstract}
A standard deep convolutional neural network paired with a suitable loss function learns compact local image descriptors that perform comparably to state-of-the art approaches.
\end{abstract} 

\section{General Learning Architecture}
Recently, several machine learning based approaches \cite{brown2010a, trzcinski2012a, simonyan2012a} have shown impressive results for finding compact low-level image representations. These representations are considered good when corresponding image patches are described by representations that are close by.

DrLim \cite{hadsell2006a} is a framework for energy based models that learns representation using only such correspondence relationships. We utilize DrLim to train a convolution neural network for learning low-dimensional mappings for low-level image patches.

The main idea behind DrLim is to map similar (i.e. corresponding) image patches to nearby points on the output manifold and dissimilar image patches to distant points. It is defined over pairs of image patches, $x_1, x_2$. The $i$-th pair $(x^i_1, x^i_2)$ is associated with a label $y^i$, with $y^i=1$ if $x^i_1$ and $x^i_2$ are deemed similar and $y^i=0$ otherwise. We denote by $d(x_1, x_2; \theta)$ the parameterized distance function between the representations of $x_1$ and $x_2$ that we want to learn. Based on $d(x_1, x_2;\theta)$ we define DrLim's loss function $\ell(\theta)$:
\[
\ell(\theta) = \sum_{i} y^i\ell_{\textrm{\tiny pll}}(d(x^i_1, x^i_2; \theta)) + (1-y^i)\ell_{\textrm{\tiny psh}}(d(x^i_1, x^i_2;\theta))
\]
We denote with $\ell_{\mathrm{\tiny pll}}(\cdot)$ the partial loss function for similar pairs (it \emph{pulls} similar pairs together) and with $\ell_{\mathrm{\tiny psh}}(\cdot)$ the partial loss function for dissimilar pairs (it \emph{pushes} dissimilar pairs apart).  $\ell_{\mathrm{\tiny psh}}$ is defined as in \cite{hadsell2006a}:
\[
\ell_{\mathrm{\tiny psh}}(d(x_1, x_2; \theta)) = 
c_{\mathrm{\tiny psh}}[\mathrm{max}
(0, m_{\mathrm{\tiny psh}} - d(x_1, x_2; \theta))]^2
\]
$m_{\mathrm{\tiny psh}}$ is the push \emph{margin}: Dissimilar pairs are not pushed farther apart if they already are at a distance greater than the push margin. $c_\mathrm{psh}$ is a scaling factor.

For $\ell_{\mathrm{\tiny pll}}$ we use a loss similar to hinge loss, differently to the loss function proposed in the original DrLim formulation:
\[
\ell_{\mathrm{\tiny pll}}(d(x_1, x_2; \theta)) = 
c_{\mathrm{\tiny pll}}[\mathrm{max}
(0, d(x_1, x_2; \theta)-m_{\mathrm{\tiny pll}})]
\]
$c_\mathrm{\tiny pll}$ is a scaling factor, $m_{\mathrm{\tiny pll}}$ is a pull \emph{margin}: Similar pairs are pulled together only if they are at a distance above $m_{\mathrm{\tiny pll}}$.

$d(x_1, x_2; \theta)$ is defined as the Euclidean distance between the learned representations of $x_1$ and $x_2$:
\[
d(x_1, x_2; \theta) = \| f(x_1; \theta) - f(x_2;\theta) \|_2
\]
$f(\cdot)$ denotes the mapping from the (high-dimensional) input space to the low-dimensional space. In this paper, $f$ is a convolutional neural network\cite{jarrett2009a}.  The layers of the convolutional network comprise a convolutional layer $C_1$ (kernel size $5 \times 5$) with 6 feature maps, a subsampling layer $S_1$, a second convolutional layer $C_2$  (kernel size $6 \times 6$) with 21 feature maps, a subsampling layer $S_2$, a third convolutional layer $C_3$ (kernel size $5\times5$) with 55 feature maps and a fully connected layer with 32 units.

\section{Experiments}
We evaluate our proposed model on the dataset from \cite{brown2010a}.
The dataset is based on more than 1.5 million image patches ($64 \times 64$ pixels) of three different scenes: the Statue of Liberty (about 450,000 patches), Notre Dame (about 450,000 patches) and Yosemite’s Half Dome (about 650,000 patches). We denote these scenes with LY, ND and HD respectively. There are 250000 corresponding image patch pairs and 250000 non-corresponding image patch pairs available for every scene. We train on one scene and evaluate the learned embedding function on the other two scenes. Evaluation is done on the same test sets (50000 matching and non-matching pairs) used also by other approaches. 

Table~\ref{tab:evs} shows that convolutional networks (last entry) perform comparably to other state-of-the-art approaches. The appeal of a simple parameteric model like a convolutional neural network is that it does not require any complex paramter tuning or pipeline optimization and that it can be integrated into larger systems that can then be trained in an end-to-end fashion \cite{hadsell2008a}.

The architecture is trained with standard gradient descent. Training stops when a local minima of the DrLim objective is reached. Notably, the hyperparameters ($c_\mathrm{\tiny pll}$, $m_{\mathrm{\tiny pll}}$, $c_\mathrm{\tiny psh}$, $m_{\mathrm{\tiny psh}}$) used in our evaluation are \emph{not} scene dependent. 

\setlength{\tabcolsep}{2pt}
\begin{tiny}
\begin{table}[ht]

   \begin{tabular}{llccc}
        \toprule
        & & \multicolumn{3}{c}{\textbf{Test set}} \\
        \textbf{Method} & \textbf{Tr. set} & LY & ND & HD \\
        \toprule
        SIFT & -- & 31.7 & 22.8 & 25.6 \\
        \\
        & LY & -- & 14.1 & 19.6 \\[-0.9ex]
        \raisebox{1.1ex}{L-BGM}& ND & 18.0 & -- & 15.8 \\[-0.9ex]
        \raisebox{1.1ex}{(64d)}& HD & 21.0 & 13.7 & -- \\

        \\
        & LY & -- & $\times$ & $\times$ \\[-0.9ex]
        \raisebox{1.1ex}{\citeauthor{brown2010a}}& ND & 16.8 & -- & 13.5 \\[-0.9ex]
        \raisebox{1.1ex}{(29d)}& HD & 18.2 & 11.9 & -- \\

        \\
        & LY & -- & $\times$ & $\times$ \\[-0.9ex]
        \raisebox{1.1ex}{\citeauthor{simonyan2012a}}& ND & 14.5 & -- & 12.5 \\[-0.9ex]
        \raisebox{1.1ex}{(29d)}& HD & 17.4 & 9.6 & -- \\

        \\
        & LY & -- & 11.2{\tiny $\pm0.3$}   & 18.5{\tiny $\pm0.5$} \\[-0.9ex]
        \raisebox{1.1ex}{CNN} & ND & 16.4{\tiny $\pm0.3$} & -- & 16.2{\tiny $\pm0.3$} \\[-0.9ex]
        \raisebox{1.1ex}{(32d)} & HD & 18.9{\tiny $\pm0.4$} & 10.7{\tiny $\pm0.2$} & -- \\
        \bottomrule
    \end{tabular}
\caption{Error rates, i.e. the percent of incorrect matches when 95\% of the true matches are found. Every subtable, indicated by an entry in the \emph{Method} column, denotes a descriptor algorithm. The line below every method denotes the size of the desciptor (e.g. 32d denotes a 32 dimensional descriptor). The 128 dimensional SIFT descriptor \cite{lowe2004a} does not require learning (denoted by $-–$ in the column \emph{Tr. set} (i.e. Training set)). The numbers in the columns labeled LY, ND and HD are the error rates of a method on the respective test set for this scene. \cite{brown2010a, simonyan2012a} do not have results when trainend on the LY scene (indicated by $\times$). L-BGM is presented in \cite{trzcinski2012a}. The mean error rates for convolutional neural networks (CNN) are given with a standard deviation over 10 runs.}
\label{tab:evs}
\end{table}
\end{tiny}

\section{More data}
Convolutional Neural Networks benefit from abundant data \cite{ciresan2012b, krizhevsky2012a}. Utilizing data from two scenes improves error rates noticebly: We get 15.1\% on LY with combined training on ND and HD (in total 1M patch pairs). Similarly, we get 8.5\% on ND and 14.3\% on HD.

%

\bibliography{drlim4lidbib}
\bibliographystyle{plainnat}

\end{document}